# Simplifying Causality: A Brief Review of Philosophical Views and Definitions with Examples from Economics, Education, Medicine, Policy, Physics and Engineering

M.Z. Naser, PhD, PE

School of Civil & Environmental Engineering and Earth Sciences (SCEEES), Clemson University, USA

Artificial Intelligence Research Institute for Science and Engineering (AIRISE), Clemson University, USA

E-mail: mznaser@clemson.edu, Website: www.mznaser.com

**Abstract**

This short paper compiles the big ideas behind some philosophical views, definitions, and examples of causality. This collection spans the realms of the four commonly adopted approaches to causality: Hume's regularity, counterfactual, manipulation, and mechanisms. This short review is motivated by presenting simplified views and definitions and then supplements them with examples from various fields, including economics, education, medicine, politics, physics, and engineering. It is the hope that this short review comes in handy for new and interested readers with little knowledge of causality and causal inference.

**Introduction**

Causality is the science of cause and effect [1]. As identifying causal mechanisms is often regarded as a fundamental purist in most sciences, causality becomes elemental in advancing our knowledge. While causality is more profound in some research areas, the concept of causality is often vague or forgotten in others [2]. With the advent of data science and the ready accessibility of big data, there is a rising potential to leverage such data in pursuit of unlocking previously unknown, hidden mechanisms or perhaps confirming ongoing hypotheses and empirical knowledge [3].

Traditionally, researchers would collect data pertaining to a phenomenon and then analyze this data to describe it, creating a model that could be used to predict such a phenomenon and/or causally infer (or explain/understand) interesting questions about such a phenomenon (see Table 1) [4].

When the primary goal is to describe the data on hand, the researcher simply aims to visualize the data to tell its story. Such visualization can take a number of fronts, i.e., examine data distribution, display and quantify the magnitude or existence of relationships and associations within the data, etc. For example, one can report statistical insights (mean, median, etc.), correlations between variables, etc. When the primary goal is to describe data, we rarely delve into how and why such relationships/associations exist.

To be able to predict, on the other hand, some variables from the data are to be identified as predictors, and at least one variable is selected as an outcome (or response). For example, one can think of regression as a mapping function wherein a list of predictors is tied together to realize an outcome. By doing so, the researcher declares the existence of a form of relationship between all variables, and such assignment can be made via assumptions and/or from domain and expert knowledge [5]. Such an assignment implicitly assumes that all data conditions have been met; hence, our view of the world does not change [4,6]. Simply put, the form of a regression equation is identical to the essence of the innings of the phenomenon at hand.

Finally, to causally infer, explain, or understand a phenomenon, the researcher ought to go beyond mere descriptions and predictions. Such a researcher would hope to uncover the data generating process (DGP) responsible for the phenomenon at hand. In some instances, one may postulate a theory with a/series of hypotheses to identify and lay out how variables can become causes or are, in fact, causes and not effects[1].

Here, a question may arise as to how a regression formula differs from a DGP? Shmueli [7] presents a brief historical review of such a question and then outlines that the type of uncertainty associated with explanations and that with predictions are different [8]. Further, a causal analysis ties a variable to a cause in support of a causal theory (DGP model), whereas a predictive analysis captures associations between a predictor and an outcome as obtained from the data at hand.

Unlike a causal model, which is retrospective, a predictive model aims to attain high accuracy in forecasting new and future observations. Thus, in the latter, a researcher focuses on selecting high quality predictors with sufficiently available and quality data rather than on the role of such predictors or the mechanics behind their relationships. By contrast, in the former, the same researcher strives to study the causal role of predictors and may, in fact, opt to retain a variable with strong causal ties despite being statistically insignificant.

Table 1 List of scientific questions

|  | Descriptive | Predictive | Causal |
|---|---|---|---|
| Medicine [4] | How can patients with specific characteristics be partitioned into groups defined by their characteristics? | What is the probability of having a particular medical condition for patients with specific characteristics? | Will procedure C reduce, on average, the risk of a particular medical condition for patients with specific characteristics? |
| Economics [3,9] | What types of businesses have a large default ratio? | What will happen to employment rates if the minimum wage is doubled? | What is the effect of a higher rate of inflation on oil prices? |
| Education [10] | Will using a new tutoring system be positively associated with improved behavioral outcomes in a given student population? | Would it be possible to identify which students are likely to respond to a structured learning environment versus medication? | What causes students to seek higher education? |
| Politics | What is the percentage of households that consistently participate in the census? | Which households are likely to vote for party P? | How can district C improve participation in its new training program? |
| Engineering | How many communities suffer from wildfires annually? | Can we predict the degree of fire-induced damage in a five-story building? | Why are buildings in neighborhood C more resilient than neighborhood B? |

Note: Some of the listed questions are inspired by their cited sources.

## The four approaches to causality – simplified[2]

Brady[3] [11] discusses four approaches to causality. These approaches include the Humean, counterfactual, manipulation, and mechanisms. Each approach defines causality in a distinct manner, and these definitions are briefly examined herein with a series of examples (including one running example). For a more in-depth review, the reader may refer to Brady [11], as well as the cited work in each section.

---

[1] For completion, Shmueli [7] shows that sometimes, in finite samples, simple models can outperform DGPs.
[2] For discussion that pre-dates the modern age of causality, please refer to [74,75].
[3] Also, inspired this brief review..

*The Humean approach*

This approach was championed by Hume [12], Mill [13], and Hempel [14], to name a few. In this approach, causality is defined as a form of regularity wherein "*X is a cause of Y if and only if X is sufficient for Y.*"

More specifically, Hume states, "*We may define a cause to be an object, followed by another, and where all the objects similar to the first are followed by objects similar to the second.*" Similarly, Mill [13] says that "*a and c are not effects of A, for they were not produced by it ...*" Finally, according to Hempel [14], "*If E describes a particular event, then the antecedent conditions Cl, C2, . . . Ck may be said to jointly 'cause' that event, in the sense that there are certain empirical regularities, expressed by the laws LI, L2,... Lr, which imply that whenever conditions of the kind indicated by Cl, C2,. .. Ck occur, an event of the kind described in E will take place*".

The Humean definition clearly assumes that a cause will always lead to the effect, and the effect will always follow the cause. Thus, causality is expected to have contiguity and succession in space and time. The following is a collection of examples from the open literature that builds on the above Humean definition of causality with commentary (whenever applicable).

Examples

*Medicine*

- Kelly et al. [15] show the following example, "*Thus, if an observational study can demonstrate that the cause always <u>precedes</u> the effect, that the effect is <u>consistently</u> close to the cause, and that the <u>association</u> is repeatedly and constantly <u>observed</u>, we can in fact still claim causation in a Humean sense.*"
- Babones [16] shows, "*... countries experiencing government turmoil are <u>likely to have</u> poor population health outcomes, as well as to be missing data.*"
  - An observational linkage between poor health conditions to movement turmoil is presented in this example.

*Economics*

- Hume [12] says, "*the <u>conjunction</u> between motives and voluntary actions is as <u>regular</u> and <u>uniform</u>, as that between the cause and effect in any part of nature.*"
  - Here, motives are seen to trigger voluntary actions, just as causes lead to effects.
- Hazlitt [17] observed, "*... when an inflation has long gone on at a certain rate, the public <u>expects</u> it to continue at that rate. More and more people's actions and demands are <u>adjusted</u> to that expectation.*"
  - This example shows how the public perceives and adjusts to common observations with regard to the state of an ongoing economy.

*Education*

- Kenny [18] says, "*A commonly cited example of spuriousness is the correlation of shoe size with verbal achievement among young children. The <u>relationship</u> is spurious because increasing age <u>causes</u> increasing shoe size and verbal achievement.*"

- o A clear example of the Humean approach to causality that ties two observations (shoe size and verbal achievement).
- Price [19] quotes Nick Gibb and mentions, "*Whatever the reason for a child's absence from school, the <u>data</u> shows that when children miss a substantial part of the school term their academic achievement <u>suffers</u> permanently.*"
  - o This example implies that observational and correlational data can be linked to causation.

## Policy

- Scoccia [20] says, "*... because Hume is a utilitarian, he would have to say that in societies where the government should be democratic, the reason why it should be democratic is that such government produces better, more just laws than a nondemocratic government would. Nondemocratic ("authoritarian") government has a <u>greater tendency</u> to become oppressive, corrupt, and inefficient than democratic government.*"
  - o This example emphasizes preconceptions of democratic and authoritarian governments in which democratic governments are seen to be more just.

## Physics and Engineering

- Rothman [21] illustrates, "*... changing the position of a light switch on the wall has the instant effect of causing the light to go on or off.*"
  - o The co-occurrence between lighting the bulb and flicking the switch can be seen to be causal in appearance. However, if the circuit is shorted, then the co-occurrence stops.
- Price [19] also notes, "*...before the Apollo Mission to the moon, in which samples of moon rocks were collected, scientists were fairly certain that there would be no sedimentary rock on the moon, since sedimentation is a process that requires the movement of materials such as water and wind, and neither of these is present on the moon due to its lack of an atmosphere. ... scientists are today confident to make highly <u>generalized</u> statements about moon rocks, their composition and the causes of their particular characteristics, despite having <u>only seen</u> rocks from three sites on the moon and despite having not actually witnessed the events that caused the rocks.*"
  - o An example with emphasis on how limited observations and knowledge from the Earthly conditions are used to make conclusive statements about the moon.
- Aged structures suffer from cracking.
  - o As expected, aging structures often suffer from structural issues such as cracking and corrosion. In essence, structural issues arise with continued use and service time.

## Running example

- From Jeffery [22] and [23]. "*Suppose that whenever the barometric pressure[4] in a certain region drops below a certain level, two things happen. First, the height of the column of mercury in a particular barometer drops below a certain level. Shortly afterwards, a storm occurs. ... Then, it may well also be the case that whenever the column of mercury drops, there will be a storm. If so, a simple <u>regularity</u> theory would seem to rule that the drop of*

---

[4] It is interesting to note that the barometer example also appears in the work of Lewis.

*the mercury column causes the storm. In fact, however, the regularity relating these two events is <u>spurious</u>. "*

- o This running example aims to first establish the view Humean approach wherein regularity gets the prime attention. Then, the same example goes on to identify that such regularity can be spurious as it bears no physical meaning.

<u>Limitations and philosophical questions</u>

This approach can be thought of as logical and deterministic, with temporal precedence. In a way, we expect an effect to occur only once its cause occurs (hence, the regularity or the law-likeness). However, an effect could occur owing to multiple causes, which leads to the issue of articulating multiple causes. Similarly, suppose regularity is the primary cause of causality. In that case, deaefferring between accidents, chance, correlations, and genuine causal effects becomes difficult (i.e., suns rise from the east, humans are mortal, etc.). Finally, the cause may require a threshold to cause this effect. This implies that even if the cause is present, its threshold may not be reached, and hence the effect may not occur.

*The counterfactual approach*

The counterfactual approach was led by Weber [24] and Simon and Rescher [25] and popularized by Lewis' [26,27] take on Hume's second definition of causality, "*if the first object had not been, the second never had existed*."

Lewis [26] defines causality as, "*We think of a cause as something that makes a difference, and the difference it makes must be a difference from what would have happened without it. Had it been absent, its effects – some of them, at least, and usually all – would have been absent as well.*" Simply put, the counterfactual approach is closely related to singular events. In this approach, the notion of "closest possible world" or a "parallel universe" parallel universe' comes to the picture of evaluating whether a counterfactual (i.e., the effect occurring without the cause) holds true. Implicitly, the counterfactual approach to causality allows us to reason without having to actually observe the cause and effect (since, by definition, a counterfactual is an event that we cannot observe).

The following is a list of counterfactual examples from the open literature[5] built on the above definition. A brief commentary has been provided.

<u>Examples</u>

*Medicine*

- Street et al. [28] argue, "*Inherent in patients' accounts of medical errors or "close calls" are <u>counterfactuals</u>. That is, when thinking about what went wrong and how it could (should) have been avoided, individuals mentally simulate <u>alternatives</u> that are more desirable than reality. This kind of "upward" counterfactual thinking includes thoughts of how a bad outcome <u>might have been</u> prevented. For example, … "The surgeon should have talked to the radiologist.".*"
  - o This example shows the idea of the closest possible world in terms of counterfactuals. In addition, it also shows how a medical could have been prevented if the responsible actor (say, a surgeon) followed a different course of action (e.g., talking to the radiologist).

---

[5] Not surprisingly, counterfactuals in the context of causality also appear in ancient history [76].

- Russo et al. [29] present: "*If an hour ago I <u>had</u> taken two aspirins instead of just a glass of water, my headache <u>would</u> now be gone.*"
  - o The counterfactual cause involves taking two aspirins to cure the headache (effect).

### *Economics*

- Heckman [30] shows that "*Counterfactuals are possible outcomes in different hypothetical states of the world. An example would be the health outcomes for a person associated with <u>taking</u> or <u>not</u> taking a drug.*"
  - o The outcome of a patient depends on the action taken by the patient ( taking or not taking a drug). Hence, one can expect the counterfactual outcome to be associated with a action that has not been taken.
- Kluve [31] says "*<u>If</u> John participated in the computer course, he <u>would</u> find a job.*"
  - o A simple example of job finding being the effect of participating in a computer course. Evidently, John does not have a job since he did not participate in the program.

### *Education*

- Holland [32] states that "*It would be of interest to a state education director who wanted to know what reading program would be the best to give to all of the first graders in his state. The <u>average causal effect</u> of the best program would be reflected in increases in statewide average reading scores.*"
  - o As we will see later, Rubin's potential outcomes rely on the difference between two states, and one is observed and the other is not (i.e., counterfactual).
- Elwert [33] presents, "*What is the <u>causal effect</u> of attending catholic school vs. public school on high school graduation?*"
  - o This implies that a causal difference exists if a student attends a Catholic or public school. Since the student can only attend one school at a time, then the counterfactual is the state of not going to the other school.

### *Policy*

- Reiss et al. [34] provide two examples:
  - o "*We might for example ask what <u>would have happened</u> to the profits of British utilities <u>had they not been</u> privatised or whether the 9/11 incidents <u>would have occurred had</u> investments in national security <u>been</u> as high as demanded by some Democrats.*"
    - ▪ This is a two-part counterfactual example in which two counterfactuals take place. In this first, a question as to how the degree of expected profit in the event that the British utilize had not been privatized. The second is the question of would larger investments in national security have prevented the 9/11 incident.
  - o A more general example with a theme on policy is, "*Policy analysts are often interested in answers to questions with the structure 'What <u>would have happened</u> to Y (the 'target variable'), <u>had X been</u> x (the 'control variable' and its value, respectively)?*"

- Levy [35] "*The First World War <u>would not have happened without</u> the assassination of Austrian archduke Franz Ferdinand*."
  - o A direct counterfactual that speculates the breakout of World War I to the assassination of Ferdinand.

*Physics and Engineering*
- Hausman [36] presents the following engineering example of a counterfactual, "*What would <u>happen</u> [to a nuclear power plant] <u>if the steam pipe were to burst?</u>*"
  - o Another direct counterfactual is that the nuclear power plant would operate normally if the steam pipe did not burst.
- Would the house <u>have burned</u> <u>had</u> it <u>not</u> been properly designed against wildfires?
  - o The poor design of the house (i.e., lack of thermal insulation) caused it to burn. Thus, if proper insulation was provided, the house would withstand the adverse effects of a wildfire.

*Running example*
- From Hannart et al. [37]. "*let R be a rainy episode and B be a downward move of the barometer's needle; then, observing R while impeding B—that is, by holding the barometer's needle—provides <u>counterfactual evidence</u> that falling barometers do not cause rain.*"
  - o This example continues our running example and showcases the counterfactual to the regularity presented by Hume. For the reader, the act of *holding the barometer's needle* can also be thought of as a manipulation.

<u>Limitations and philosophical questions</u>
Counterfactual could be perceived in a literal sense, wherein a cause also becomes a necessary condition for the effect to occur. Kelly et al. [15] reinforce the idea of overdetermination by noting, "*… if only necessary conditions are causes, then if there are two causes that are each sufficient for the effect, then neither is necessary, thus neither is a cause*." For example, revisiting our last example, one could debate that if the house were not built in the first place, then it would not have burned down. Another issue within the counterfactual approach revolves around handling multiple causes and how to actually prove the counterfactual in the closest possible world. Finally, Reiss et al. [34] stipulate the use of counterfactuals in policy research is to describe what would happen once a policy is put into place, as opposed to articulating its effects.

*The manipulation approach*
The manipulation approach appears in the works of a number of philosophers, such as Gasking [38], Sobel [39], Heckman [40], Cartwright [41], and Menzies and Price [42]. This approach stems from creating a blueprint (or recipe) that continually produces an effect from the cause.

For example, Gasking defines the manipulation approach as [38] "*the notion of causation is essentially connected with our manipulative techniques for producing results*." Similarly, for Sobel [39] "*view the cause as an event or state that can be manipulated (or at least potentially manipulated); under this view, causation resides in the existence of a one-to-one correspondence between the state of the manipulated cause and the state of the effect*." And Heckman [40] states "*Holding all factors save one at a constant level, the change in the outcome associated with manipulation of the varied factor is called a causal effect of the manipulated factor.*"

Cartwright [41] takes the position of "*The manipulation view of causation revolves around the idea that causes give us effective strategies for producing effects we want, or preventing those we do not; by manipulating the cause we can manipulate the effect in a predictable way.*" Finally, Menzies and Price [42] define manipulation as "*events are causally related just in case the situation involving them possesses intrinsic features that either support a meansend relation between the events as is, or are identical with (or closely similar to) those of another situation involving an analogous pair of means-end related events.*"

Here are some examples that revolve around the manipulation approach, with a brief commentary.

<u>Examples</u>

*Medicine*

- Nolan and O'Connor [43] illustrate, "*The observation that manipulating causal attributions can have an immediate effect on treatment preferences also raises the possibility that clinicians could seek to <u>modulate</u> patients' treatment preferences by <u>strategically</u> providing aetiological information.*"
  - o In other words, once a recipe for manipulation is arrived at, then this recipe could be used to tailor treatments in such a way as to yield specific effects.
- Mirtz et al. [44] show, "*From the therapeutic viewpoint if a person is said to have subluxation then a <u>certain</u> amount of spinal <u>manipulation</u> <u>should</u> show a response, in some form, by the human body.*"
  - o Similar to the above example, there is an expectation that some form of response is tied to the act of manipulation.

*Economics*

- Cartwright [41] states that "*If, though, the government <u>manipulates</u> inflation to improve unemployment, entrepreneurs will recognize the rise in prices for what it is—just inflation—and will not open new jobs.*"
  - o This implies that the government already has a way (i.e., a means/recipe) to manipulate inflation successfully. Thus, inflation could be tweaked to the degree required by the government.
- Peters [45] says, "*The most obvious example would be field <u>experiments</u> in which the researcher, or a government, <u>manipulates</u> an important variable and then <u>assesses</u> the outcomes. For example, Finland conducted an experiment in income maintenance policy in which a sample of unemployed were given an income that would not be reduced if they also found work.*"
  - o The mention of experiments denotes the ability to strategically manipulate a predefined set of conditions and variables.

*Education*

- Cartwright [41] states that "<u>*Reducing*</u> *class sizes* <u>*causes improved*</u> *reading scores.*"
  - o The verb, reducing, acts as a surrogate for manipulating the size of a given class to examine the result of this action.
- Introducing a new program (e.g., training teachers) is often considered a manipulation exercise.

- o This example parallels the example of Cartwright. The introduction of a training program is a manipulation of the existing level of teacher skills.

*Policy*

- Duncan and Magnuson [46] state, "*Income is the component of SES [Socioeconomic status] that is most <u>directly manipulated</u> by policy.*"
  - o In other words, policies, when applied in the context of SES, can directly manipulate income.
- Contreary et al. [47] say, *"... beneficiaries might be able to influence their exposure to a policy by <u>manipulating</u> their score on the eligibility variable— for example, by misreporting their income.*"
  - o The action of misreporting income denotes manipulating eligibility scores (in the hope of gaining benefits).

*Physics and Engineering*

- Blanchard [48] presents, "*Paradigmatic examples of causes are local events: the <u>throwing</u> of the rock <u>causing</u> the window to break, the <u>lighting</u> of the match <u>causing</u> the forest to burn, etc.*"
  - o The interventions (or manipulations) on a given state are noted by the verbs of *throwing* and *lighting*.
- Altering the loading setup from flexural to shear would cause a loaded beam to fail in shear.
  - o A simple engineering example wherein changing an experimental setup highly influences the outcome of the experiment.

*Running example*

- From Maher [49]. "*The reading on a barometer is not a cause of rain because we cannot <u>change</u> whether it rains by <u>manipulating</u> the barometer reading.*"
  - o This is a clear example of how changing the position of the barometer's needle does not affect the chances of rain.

<u>Limitations and philosophical questions</u>

Perhaps one of the key limitations to manipulation arises from what Brady [11] states as a strength, "*The combination of intervention and control in experiments makes them especially effective ways to identify causal relationships.*" This is because the heavy reliance on experimentation and physical intervention may hinder establishing causality when these two actions are not possible[6]. Similarly, in an experiment, the cause is typically manipulated. If the cause is unidentifiable or mistakenly identifiable, the effect may not be realized accurately. A similar argument can be made to accurately apply the degree of manipulation or robust measuring/declaring its effect (e.g., via specialized or sensitive sensors). Finally, Bullock et al. [50] remind us that "*a common criticism of experiments is that they reveal but do not explain causal relationships.*" The companion work of Spirtes and Scheines [51] further complements this view.

---

[6] Say, being not ethical or resource intensive – which moves this discussion towards the realm of quasi-experimentation (see [77]). May also arise from the lack of equipment or expertise needed to carry out such manipulation.

*The mechanism approach*

One of the simplest ways to think about mechanisms is to think about processes (i.e., biological or chemical processes, etc.) that have an inherent temporal order. For example, Machamer [52] defines mechanisms as "*entities and activities organized such that they are productive of regular changes from start or setup to finish or termination conditions*." Glennan [53] also defines mechanisms as "*complex systems whose "internal" parts interact to produce a system's "external" behavior. I argue that all but the fundamental laws of physics can be explained by reference to mechanisms*." A more direct definition of mechanisms is articulated by Gillies [54], "*A causes B if and only if there is a mechanism which links A to B*."

Herein, we present a list of examples that reinforce the mechanistic approach to causality.

Examples

*Medicine*

- Williamson [55] presents, "*An example is the mechanism by which the heart pumps blood*."
  - Simply, the pumping of the blood follows a process (or a mechanism) that is thoroughly examined in medicine.
- Kendall-Tackett and Klest [56] note, "*More recently, they have recognized that negative mental states are also mechanisms that can lead to poor health, even increasing the risk of premature mortality*."
  - Poor health and premature mortality stem from the mechanism of negative mental states.

*Economics*

- Ridley et al. [57] explain "… *depressed individuals might avoid making active choices and may stick with "default options," may have decreased sensitivity to incentives because of anhedonia, or may have difficulty choosing among several options. Understanding the importance of this mechanism relative to more "direct" economic effects through disability or health expenditures is crucial for correctly measuring the economic burden of mental illness and designing economic policy for those whose mental health is compromised*."
  - This example shows that understanding the responsible mechanism opens the door to capturing economic burdens.
- Rothstein [58] notes, "*Instead, I am trying to make sense of how the causal mechanism between trust in political institutions and 'generalized trust' may work. One could argue, of course, that the causal mechanism runs the other way—that people known to be trustworthy would not try to corrupt government officials*."
  - In this example, the same mechanism can be explained by multiple views, thus implying a less deterministic nature of the mechanisms.

*Education*

- Kawachi et al. note [59], "*The causal mechanisms underlying the link between schooling and health may operate through both material and psychosocial mechanisms. Education equips individuals with general as well as specific knowledge and skills that are useful for prevention of disease*."
  - Education serves as the primary carrier of a mechanism to prevent diseases.

- Bjorklund and Salvanes [60] say, "*... there are two reasons or <u>mechanisms</u> for <u>a positive correlation in schooling</u> across generations: 1) abilities are correlated across generations and they raise schooling for both parents and children; and 2) parents' human capital has a direct effect on their children's schooling because it directly affects how productive the children are in terms of schooling.*"
  - o This example describes an interesting equality between reasons, correlations, and correlations. Possibly, the observed correlations can be stipulated to be driven by an underlying mechanism.

*Policy*

- Falleti and Lynch [61] and King et al. [62] show, "*For example, variables such as minority disaffection and governmental decisiveness are the <u>mechanisms</u> that <u>explain</u> how the political system (presidential or parliamentary) affects democratic stability in a hypothetical large-sample research study.*"
  - o This example illustrates a direct link between explanations and mechanisms in the context of democratic stability.
- Capano and Howlett [63] say, "*In the policy realm, these <u>mechanisms</u> exist at the individual level whereby actions like the provision of subsidies are expected to change individual savings behaviour...*"

*Physics and Engineering*

- Glennan [53] presents the following example, "*Consider a system consisting of a series of three or more gears of various sizes. Given information about the number of teeth on the gears, one can state a <u>law</u> L1 <u>describing</u> the rotation of the last gear as a <u>function</u> of the rotation of the first gear.*"
  - o This example clearly presents the physical mechanism (i.e., law) for the rotation of gears.
- When steel is exposed to an acidic environment and moisture, iron is removed and dissolved in the surrounding solution.
  - o This example shows the corrosion process (or mechanism) in the presence of moisture and acids.

*Running example*

- Revisiting the barometer example from a mechanismic view requires us to arrive at an understanding of how mercury responds to atmospheric pressure.

<u>Limitations and philosophical questions</u>

One of the limitations of these mechanisms is the effective articulation of recipes. For example, a researcher may unknowingly combine a series of entities into one mechanism, wherein they ought to be considered as separate entities. The same could be true in the event that an oversimplified mechanism is sought, wherein such a mechanism is only a surrogate for parallel entities (or causes). A clear examination of these mechanisms may become cumbersome, especially when domain knowledge is limited. Brady and other philosophers point out the pairing problem, which states, "*The general problem is that constant conjunction of events is not enough to "pair-up" particular events even when preemption is not present.*"

Finally, Capano and Howlett [64] and van der Heijden [65] point out two types of mechanisms, first-order and second-order as "*Following these authors, first-order mechanisms are set in motion by such activators and "affect the behaviour of individuals, groups and structures [...] to achieve a specific outcome". Second-order mechanisms "inform the use of activators by observation of the reaction of individual, group and system behavior to the previous deployment of activators.*"

**General views on establishing causality**

Four primary views are discussed herein: Hill's [66] criteria, Cox and Wermuth's [67] notions of statistical causality, Rubin's potential outcomes, and Pearl's [68] three rungs of causality.

*Hill's criteria*

Hill [66] proposes nine criteria to determine causation, namely, strength (strong association), consistency (regularity/repeated observations), specificity (a cause leads to a single effect), temporality (a cause precedes its effect in time), biological gradient (presence of a unidirectional dose-related response), plausibility, coherence (interpretations of a cause and effect does not conflict with what is known), experiment, and analogy. These criteria were originally established in the field of epidemiology and were further generalized from there.

*Cox and Wermuth's statistical causality*

Cox and Wermuth [67] base statistical causality on three notions. These notions start with the zero-level which represents statistical associations (i.e., dependence). Such an association cannot be explained via the existing data or via additional predictors. The zero-level is then followed by the first-level. It is at this level that outcomes from interventions can be compared.

Often, a scenario is constructed in which the state of a unit (e.g., a patient) is compared under two conditions (or notational responses, as per Cox and Wermuth [67]). For example, say that a new intervention is available and a researcher is attempting to gauge the effect of such intervention versus a control (i.e., no intervention) – while other things being equal. The thought process is that an effect exists if the outcome observed from the intervention is declared to be meaningful.

Practically, only one state can be observed (i.e., the intervention or control) because, in reality, a patient can receive either an operation or a drug. This effectively implies that the unobserved outcome can be labeled as a counterfactual. While this limits our estimation of possible causal effects at the unit level, such an estimation can be aggregated at the group level to yield the average causal effect. It is through a comparison of states that causal evidence can be established.

Once findings from zero- or first-level notions have been analyzed and established, the researcher shifts gears to explain how such notions arose. Here, we can leverage a meta-analysis of various studies or carefully designed experiments to realize explanations. It should be noted that in the physical sciences, an experiment with multiple units (e.g., load-bearing members) could be designed and carried out to thoroughly examine the effect of interventions (e.g., the use of a new construction method or material).

Cox and Wermuth [67] summarize the limitations of their notions of statistical causality as follows:

- "*... zero-level causality suffers from the criticism that their is no intervention involved to observe the causal effect of doing something on the system.*"

- "*... first-level causal estimation that mostly involves randomization experiments may make the conclusions of the study more secure, but fails to reveal the ... processes working behind the effect observed.*"
- "*The second-level of causality requires field knowledge and cannot be solely data driven.*"

*Rubin's potential outcomes*

Rubin's [69] potential outcomes view is similar to the first-level notion mentioned earlier. This approach builds upon the works of Fisher [70] (who emphasized randomization) and Neyman [71]. Simply put, the effect of a cause is the difference between two states, wherein one is observed, and the other is counterfactual. For brevity, this view is not re-iterated again. However, I would strongly suggest reviewing the notable work of Rubin [72,73] (and key elements such as the Stable Unit Treatment Value Assumption (SUTVA), propensity score, etc.) as well as the extensive review and discussion by Holland [32].

*Pearl's three rungs of causality*

Pearl's [1] three rungs of causality stem from the pursuit of creating strong AI (machines with the ability to learn and make decisions like humans). These rungs can be visualized as a ladder. The first level of this ladder builds on associations. Associations are the insights we observe (i.e., data). Such associations are similar to the zero-level notion of Cox and Wermuth [67] and rely heavily on correlations.

The second level examines the role of interventions. An intervention can be considered as setting (or changing) the value of a given variable to a specific value. In a way, an intervention describes the distribution of the outcome conditional on *setting* the value of a variable, $X = x$. This distribution is distinct from the observational distribution associated with the outcome that we naturally *observe* when $X = x$.[7] As one can see, interventions are a means for active measures, and hence, they are set above associations (means from observations). For example, both physical and randomized experiments fall under the intervention category. Pearl also created *do*-calculus and a neat graphical method to describe causal relations and evaluate interventions (see [68]).

The highest rung on Pearl's ladder belongs to counterfactuals. Counterfactuals intuitively imply, counter to the fact, and hence require acts of imagination as they are elemental to inferring causes and effects. While in some fields, conducting experiments may not be possible (i.e., changing past events) nor ethical (e.g., exposing patients to questionable treatments), in others, counterfactuals could be examined via physical experiments. For instance, a load-bearing specimen such as a column can only be tested once before it collapses. Fortunately, it is possible to fabricate identical column specimens to the first column. These specimens can be tested under different conditions to examine the causal effect between such conditions (i.e., environmental exposure) and that used in the benchmark.

---

[7] In Pearl's words:

"*The association layer is characterized by conditional probability sentences, e.g., $P(y|x) = p$ stating that: the probability of event $Y = y$ given that we observed event $X = x$ is equal to $p$.*"

"*At the interventional layer we find sentences of the type $P(y|do(x), z)$, which denotes "The probability of event $Y = y$ given that we intervene and set the value of $X$ to $x$ and subsequently observe event $Z = z$.*"

"*At the counterfactual level, we have expressions of the type $P(y|x', y')$ which stand for "The probability that event $Y = y$ would be observed had $X$ been $x$, given that we actually observed $X$ to be $x'$ and $Y$ to be $y'$.*"

## Conclusions
This paper presents some of the big ideas behind key philosophical views and definitions of causality and then supports such views with examples. More specifically, four commonly adopted approaches to causality: Hume's regularity, counterfactual, manipulation, and mechanisms are briefly reviewed. The applicability of these approaches in practical scenarios, including economics, education, medicine, politics, physics, and engineering, is discussed. At a personal level, the author shares the view of Brady [11] in which "*For our purposes, we embrace them all [approaches of causality]... Therefore, practical researchers can profit from drawing lessons from each one of them even though their proponents sometimes treat them as competing or even contradictory.*"

Admittedly, this paper only presents some of the big ideas in each approach and avoids ongoing philosophical debates to simplify such ideas for new readers interested in this area. This author hopes that such readers find interest in navigating many of the aforenoted debates and lively discussions on other concepts (i.e., sufficiency, necessity, asymmetry of causation, pairing, multiple, probabilistic, and reverse causation, as well as causal preemption (i.e., redundant causes), etc.) in the pages of the cited works in this paper.

## Conflict of Interest
The author declares no conflict of interest.